\documentclass{article}

\PassOptionsToPackage{numbers, compress}{natbib}

\usepackage[final]{neurips_2021}
\usepackage[utf8]{inputenc} 
\usepackage[T1]{fontenc}    
\usepackage{url}            
\usepackage{booktabs}       
\usepackage{amsfonts}       
\usepackage{nicefrac}       
\usepackage{microtype}      
\usepackage{xcolor}         
\usepackage{bm}
\usepackage{amsmath}
\usepackage{mathtools}

\usepackage{algorithm}
\usepackage[noend]{algpseudocode}
\algnewcommand\algorithmicinput{\textbf{Input:}}
\algnewcommand\algorithmicoutput{\textbf{Output:}}
\algnewcommand\Input{\item[\algorithmicinput]}%
\algnewcommand\Output{\item[\algorithmicoutput]}%

\DeclareMathOperator*{\argmax}{arg\,max}

\newcommand{\dbf}{\mathbf{d}}
\newcommand{\ybf}{\mathbf{y}}
\newcommand{\thetab}{\bm{\theta}}
\newcommand{\psib}{\bm{\psi}}

\title{Bayesian Optimal Experimental Design for \\ Simulator Models of Cognition}

\author{%
  Simon Valentin\thanks{Equal contribution. Correspondence to:~\texttt{s.valentin@ed.ac.uk}.\vspace{-0.3cm}} \\
  University of Edinburgh\\
   \And
  Steven Kleinegesse\footnotemark[1] \\
  University of Edinburgh \\
   \AND
  Neil R. Bramley \\
  University of Edinburgh \\
  \And
  Michael U. Gutmann \\
  University of Edinburgh \\
  \And
  Christopher G. Lucas \\
  University of Edinburgh \\
}

\begin{document}

\maketitle 

\vspace{-0.2cm}

\begin{abstract}
Bayesian optimal experimental design (BOED) is a methodology to identify experiments that are expected to yield informative data. 
Recent work in cognitive science considered BOED for computational models of human behavior with tractable and known likelihood functions.
However, tractability often comes at the cost of realism; simulator models that can capture the richness of human behavior are often intractable. 
In this work, we combine recent advances in BOED and approximate inference for intractable models, using machine-learning methods to find optimal experimental designs, approximate sufficient summary statistics and amortized posterior distributions. 
Our simulation experiments on multi-armed bandit tasks show that our method results in improved model discrimination and parameter estimation, as compared to experimental designs commonly used in the literature. 
\end{abstract}

\section{Introduction}
Computational models provide a means to describe and study natural phenomena, with important applications ranging from understanding the spread of viruses~\citep{currie2020} to discovering new molecules~\citep{jumper2021highly} and studying climate change~\cite{runge2019inferring}.
A particular scientific domain where computational models are playing an increasingly prominent role is the study of human behavior. 
Here, computational models allow us to formalize theories about human cognition and thereby better understand and explain behavior.
Often, models that are rich enough to capture realistic behavior have intractable likelihood functions, complicating common scientific tasks such as model comparison, parameter inference and designing informative experiments.

Gathering experimental data is generally costly and time-consuming.
Meanwhile, behavioral experiments are typically designed based on prior work, intuitions and heuristics, which may yield data that are poor for resolving the researchers' theoretical questions. 
In particular, as psychological theories become more complex, designing informative experiments can be a difficult task. 
 
As a potential solution to this problem, the field of Bayesian optimal experimental design (BOED) formalizes the design of experiments and treats it as an optimization problem. 
Concretely, the aim is to maximize a utility function that captures the worth of a particular experimental design. 
This utility function, however, usually depends on the posterior distribution and is therefore intractable for all but the simplest heuristic models of cognition.

Prior work in cognitive science has demonstrated the applicability and usefulness of BOED for parameter estimation and model comparison, albeit for simple models with known and tractable likelihood functions~\citep[e.g.,][]{myung_optimal_2009, zhang_optimal_2010, ouyang_webppl-oed_2018}. 
There is, however, a lack of work that considers more realistic cognitive models in the context of BOED.

Nonetheless, in the recent years there have been significant methodological advances in BOED for simulator models~\citep[e.g.][]{foster_variational_2019, kleinegesse_bayesian_2020, overstall2020}.
We here leverage these advances and utilize the MINEBED method of
\citet{kleinegesse_bayesian_2020}. 
This method performs BOED by maximizing a lower bound on the expected information gain at a particular experimental design. 
This lower bound is estimated by training a neural network on data generated by the computational models under consideration.

\begin{figure}[!t]
    \centering
    \includegraphics[width=0.9\linewidth]{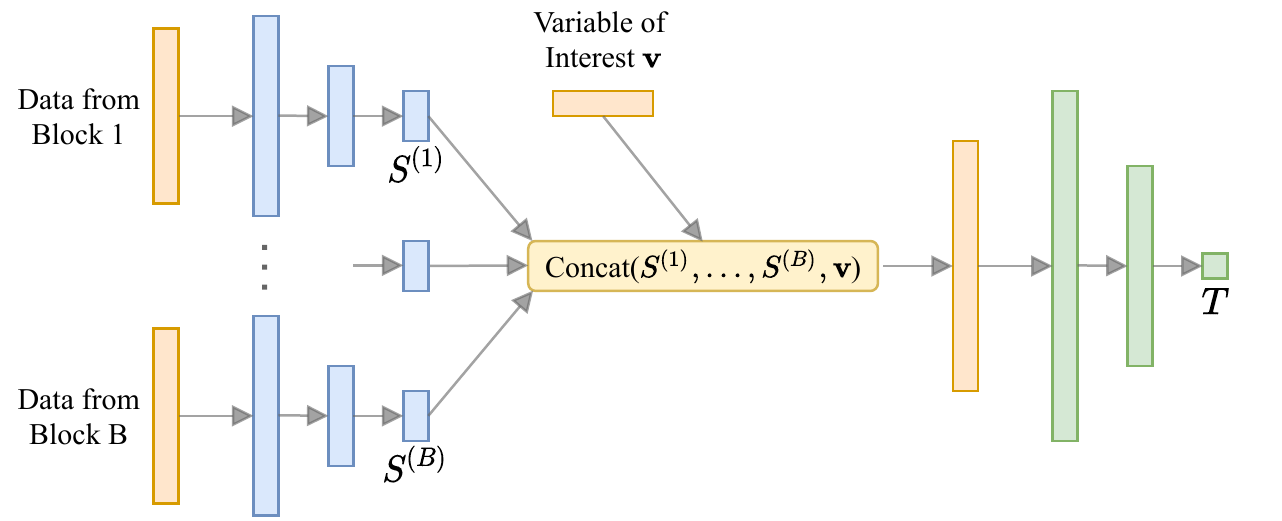}
    \caption{Neural network architecture for behavioral experiments. For each block of data we have a small sub-network (shown in blue) that outputs summary statistics $S$. These are concatenated with the variable of interest $\bm{v}$ and passed to a larger neural network (shown in green).}
    \label{fig:my_label}
\end{figure}

\paragraph{Contributions} We demonstrate the applicability of modern machine learning and BOED methods to any computational models of cognition from which we can simulate data.
To deal with data common in behavioral experiments, we construct a bespoke neural network architecture, visualized in Figure~\ref{fig:my_label}, that is trained on simulated data from our computational models. 
In addition to finding optimal designs and amortized posterior distributions, this allows us to extract approximate sufficient statistics at the same time.
As a case study, we extend previously studied models of behavior in multi-armed bandit tasks to be more flexible and realistic, which results in intractable likelihood functions. 
In our simulation study, we demonstrate how to find optimal experimental designs for model discrimination and parameter estimation.
We validate our results by comparing the performance of optimal and commonly-used experimental designs in statistical inference.

\section{Models of Human Behavior in Bandit Tasks}

We turn to models of human sequential decision making in multi-armed bandit tasks as an important area of research in cognitive science~\citep[e.g.,][]{steyvers_bayesian_2009}.
Multi-armed bandits have a long history in statistics and machine learning~\cite[e.g.,][]{robbins_aspects_1952, sutton_reinforcement_2018, steyvers_bayesian_2009} and formalize a common class of decision problems --- repeatedly choosing between a set of options under uncertainty.
Here, we select three computational models that have previously been proposed as accounts of human choice behavior. 
We generalize them to accommodate richer and more realistic behavioral patterns, which results in intractable likelihoods.

Each of our computational models defines a generative model $\mathbf{y} \sim p(\mathbf{y} | \bm{\theta}, \mathbf{d})$,
where the observed data $\mathbf{y}$ consist of the chosen bandit arms and received rewards in a particular bandit task.
The model parameters $\bm{\theta}$ are treated as random variables with known prior distributions.
The design vector $\mathbf{d}$ that we wish to optimize consists of the reward probabilities of the Bernoulli distributions associated with the arms in the bandit task.
We provide high-level intuitions for our computational models below, and give detailed explanations in the Appendix.

\paragraph{Win-Stay Lose-Thompson-Sample (WSLTS)} 
Here, we propose Win-Stay Lose-Thompson-Sample (WSLTS) as an amalgamation of Win-Stay Lose-Shift~\citep[WSLS;][]{robbins_aspects_1952} and Thompson Sampling~\citep{thompson1933likelihood}. 
Upon observing a reward, the agent re-selects the previously chosen arm with a certain probability, and shifts to another arm if there was no reward (with a different probability). 
Here, the agent performs Thompson Sampling from a reshaped posterior instead of shifting to another arm uniformly at random, as would be the case in standard WSLS.

\paragraph{Auto-regressive $\varepsilon$-Greedy (AEG)}
$\varepsilon$-Greedy~\citep[e.g.,][]{sutton_reinforcement_2018} is a ubiquitous method in reinforcement learning in which the agent selects the arm with the highest expected reward with $p = 1-\epsilon$ and a uniformly selected arm otherwise.
Here, we propose Auto-regressive $\varepsilon$-Greedy (AEG) as a generalization of
$\varepsilon$-Greedy, where the probability of selecting the previous arm is controlled by a separate parameter.
This allows for modeling people's tendency towards auto-regressive behavior~\citep[][]{schulz2020finding}.

\paragraph{Generalized Latent State (GLS)}
\citet{lee_psychological_2011} proposed a latent state model for bandit tasks whereby a learner can be in either an \emph{explore} or an \emph{exploit} state and switch between these as they go through the task. 
Here, we propose the Generalized Latent State (GLS) model, which unifies and extends latent-state and latent-switching models,  previously studied in \citet{lee_psychological_2011}, allowing for more flexible and structured transitions (see the Appendix for more details).

\section{Methods}

\paragraph{BOED Background}
In BOED, we need to construct a utility function $U(\mathbf{d})$ that describes the worth of an experimental design $\mathbf{d}$. 
Finding an optimal design $\mathbf{d}^{\ast}$ then equates to maximizing this utility function, i.e.~$\mathbf{d}^{\ast} = \argmax_{\mathbf{d}} U(\mathbf{d})$. 
A prominent and principled choice of utility function 
is the \emph{mutual information} (MI), which is equivalent to the expected information gain, 
\begin{equation} \label{eq:mi}
U(\mathbf{d}) = \text{MI}(\bm{v}; \mathbf{y} | \mathbf{d}) \coloneqq \mathbb{E}_{p(\mathbf{y}|\bm{v}, \mathbf{d})p(\bm{v})} \left[
\log{\frac{p(\bm{v}|\mathbf{y}, \mathbf{d})}{p(\bm{v})}}\right],
\end{equation}

where $\bm{v}$ is a variable of interest that we wish to estimate.

Intuitively, mutual information quantifies the amount of information our experiment is expected to provide about the variables of interest. 
Unfortunately, computing the MI exactly is generally intractable, a difficulty that is exacerbated for intractable models. Below, we describe how the MI can be estimated and optimized effectively, by using the MINEBED methodology of~\citet{kleinegesse_bayesian_2020}. More details on the training procedure can be found in the Appendix.

\paragraph{Variable of Interest}
In this work we focus on two common scientific goals: (1) the task of model discrimination (MD), i.e.~distinguishing between competing cognitive models, and (2) the task of parameter estimation (PE) of a given cognitive model. 
For MD, the variable of interest $\bm{v}$ in Equation~\ref{eq:mi} is a discrete model indicator $m$ that determines from which competing model the data originates. 
For PE, the variable of interest is the set of parameters $\bm{\theta}_m$ of a particular model $m$.

\paragraph{MI Lower Bound} 
We are ultimately interested in maximizing the MI with respect to the designs $\mathbf{d}$, not estimating it accurately everywhere in the design domain. 
Recent advances in BOED for simulator models thus advocate the use of cheaper MI lower bounds instead~\citep[e.g.][]{foster_variational_2019,kleinegesse_bayesian_2020}.
We shall here use the MINEBED method of~\citet{kleinegesse_bayesian_2020}, which works by training a neural network $T_{\psib}(\bm{v}, \ybf)$ using stochastic gradient-ascent, where $\psib$ are the neural network parameters and the data $\mathbf{y}$ is simulated at design $\mathbf{d}$ with samples from the prior $p(\bm{v})$.
 
In particular, the neural network is trained by using the MI lower bound as an objective function (see the Appendix for the exact form of the lower bound).

The resulting trained neural network and the final lower bound estimate can thus be used to compute an estimate of the MI at design $\mathbf{d}$.

\paragraph{Network Architecture}

We propose an effective architecture choice for the neural network $T_{\psib}(\bm{v}, \ybf)$, devised specifically with applications to behavioral experiments in mind (but may also be effective in other applications).
Our architecture, summarized in Figure~\ref{fig:my_label}, incorporates sub-networks $\mathbf{S}_{\bm{\phi}}(\bm{v}, \ybf)$ for each block of behavioral data in a bandit task. 
The outputs of these sub-networks are then concatenated and passed as input to the network $T_{\psib}(\bm{v}, \ybf)$. 
Following~\citet{Chen2021a}, each sub-network is learning approximate sufficient statistics of the data from a particular block in the bandit task.

\paragraph{Gradient-Free Optimization} The MINEBED method originally optimizes the utility function $U(\mathbf{d})$ by means of gradient-ascent. 
This is however not possible when the simulated data $\mathbf{y}$ is discrete, as is commonly the case with behavioral data. 
We thus follow the third experiment in~\citet{kleinegesse_bayesian_2020} and optimize $U(\mathbf{d})$ with respect to $\mathbf{d}$ using Bayesian Optimization (BO).
We use a Gaussian Process (GP) as our probabilistic surrogate model with a Mat{\'e}rn-5/2 kernel and Expected Improvement as the acquisition function (these are standard choices, see~\citet{shahriari_taking_2015} for a review on BO).

\section{Experiments}
In this section we demonstrate the optimization of reward probabilities for multi-armed bandit tasks, with the scientific goals of (1) model discrimination (MD) and (2) parameter estimation (PE). 
We consider multi-armed bandits with three arms and $30$ trials per block, where each block may have different reward probabilities. 
For the MD task we use two blocks of behavioral data, whereas we use three blocks for the PE task. 
As part of our methodology, we use relatively small architectures with $2$ hidden layers for all neural networks. 
More information about the experimental settings, descriptions of the algorithms and additional discussions can be found in the Appendix.

\paragraph{Baseline design}
As our baseline designs, we sample all reward probabilities from a $\text{Beta}(2,2)$ distribution, following the, to the best of our knowledge, largest behavioral experiments on bandit problems in the literature with 451 participants~\citep{steyvers_bayesian_2009}.

\begin{figure}[!t]
    \centering
    \begin{minipage}{0.38\textwidth}
        \centering
        \includegraphics[width=1\linewidth]{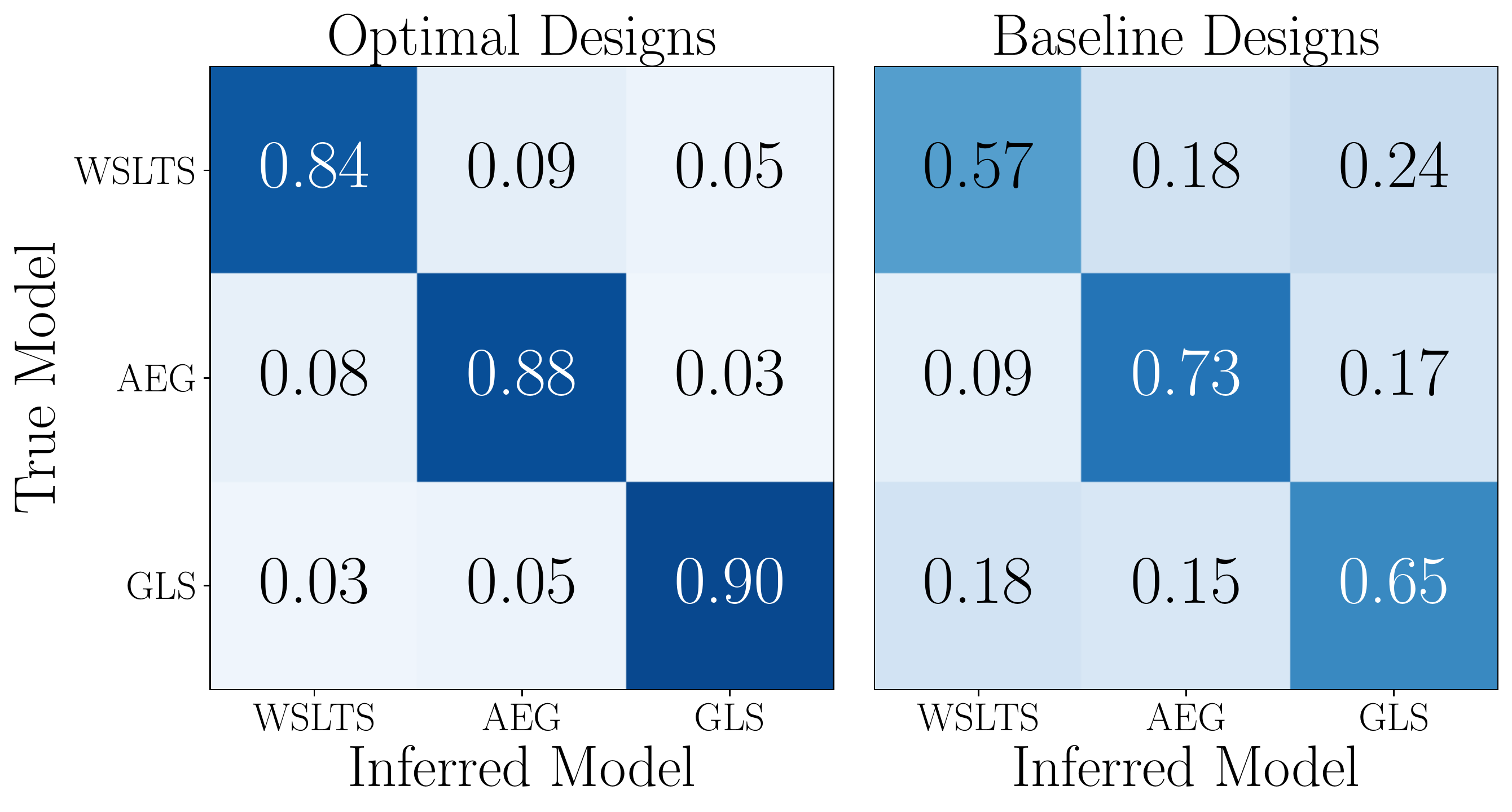}
        \caption{Confusion matrices of the inferred behavioral models, for optimal (left) and baseline (right) designs.}
        \label{fig:md_post}
    \end{minipage}%
    \hfill
    \begin{minipage}{0.58\textwidth}
        \centering
    \vspace{0.15cm}
    \includegraphics[width=1\linewidth]{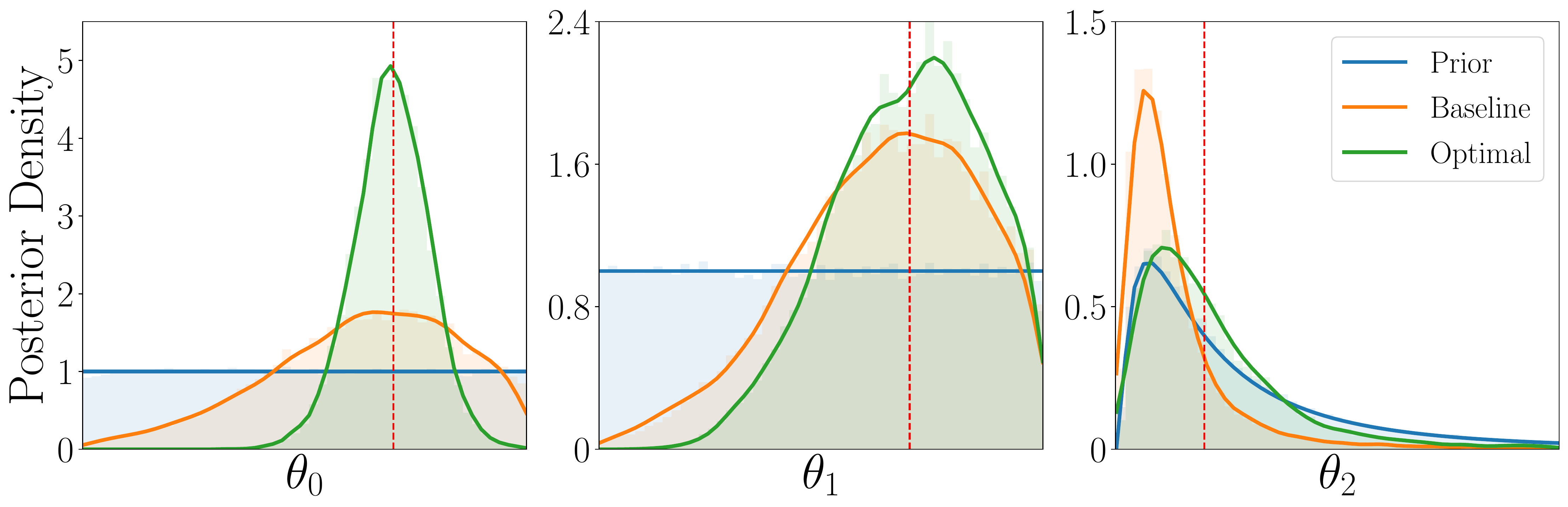}
    \vspace{0.1cm}
        \caption{Marginal posterior distributions of the three WSLTS model parameters for optimal (green) and baseline (orange) designs, averaged over $1{,}000$ observations.}
        \label{fig:pe_post}
    \end{minipage}
\end{figure}

\paragraph{Model discrimination} 
The optimal reward probabilities for the MD task found using our methodology are $[0, 0, 0.6]$ for the first block of trials and $[1, 1, 0]$ for the other second block. These optimal designs stand in stark contrast to the usual reward probabilities used in such behavioral experiments, which almost always take non-extreme values~\citep{steyvers_bayesian_2009}. 
Using the neural network that was trained at that optimal design, we can compute posterior distributions of the model indicator (as described in the appendix). This yields the confusion matrices in Figure~\ref{fig:md_post}, which show that our optimal designs result in considerably better model recovery than the baseline designs.

\paragraph{Parameter estimation} We here discuss the PE results for the WSLTS model; the results for the AEG and GLS model can be found in the appendix. We find that the optimal reward probabilities for the WSLTS model are $[0, 1, 0]$, $[0, 1, 1]$ and $[1, 0, 1]$ for the first, second and third block, respectively. 
Similarly to the MD task, these optimal designs take extreme values, unlike commonly-used reward probabilities in the literature. 
In Figure~\ref{fig:pe_post} we show posterior distributions of the WSLTS model parameters for optimal and baseline designs. 
We find that optimal designs yield data that result in considerably improved parameter recovery.

\section{Conclusions}
Our experiments demonstrate that our methodology allows us to effectively design optimal experiments in cognitive science, resulting in considerably better model and parameter recovery than for designs commonly-used in the literature. 
In particular, the combination of a lower bound on the MI, parameterized by a bespoke neural network, allows us to scale to realistic behavioral experiments. 
It would be interesting to see how our approach can be adapted to the sequential BOED setting with intractable models~\citep[e.g.,][]{kleinegesse2020sequential}. 
Additionally, our experiments only included synthetic data and no real-world data, and therefore it would be useful to apply our approach to real participants as well.

\clearpage

\medskip

\section*{Acknowledgments}
SV was supported by a Principal's Career Development Scholarship,
awarded by the University of Edinburgh.
SK was supported in part by the EPSRC Centre for Doctoral Training in Data Science, funded by the UK Engineering and Physical Sciences Research Council (grant EP/L016427/1) and the University of Edinburgh.

{
\small

\bibliography{references.bib}

\begin{thebibliography}{21}
\providecommand{\natexlab}[1]{#1}
\providecommand{\url}[1]{\texttt{#1}}
\expandafter\ifx\csname urlstyle\endcsname\relax
  \providecommand{\doi}[1]{doi: #1}\else
  \providecommand{\doi}{doi: \begingroup \urlstyle{rm}\Url}\fi

\bibitem[Chapelle and Li(2011)]{chapelle2011empirical}
O.~Chapelle and L.~Li.
\newblock An empirical evaluation of thompson sampling.
\newblock \emph{Advances in neural information processing systems},
  24:\penalty0 2249--2257, 2011.

\bibitem[Chen et~al.(2021)Chen, Zhang, Gutmann, Courville, and Zhu]{Chen2021a}
Y.~Chen, D.~Zhang, M.~U. Gutmann, A.~Courville, and Z.~Zhu.
\newblock Neural approximate sufficient statistics for implicit models.
\newblock In \emph{International Conference on Learning Representations
  (ICLR)}, 2021.

\bibitem[Currie et~al.(2020)Currie, Fowler, Kotiadis, Monks, Onggo, Robertson,
  and Tako]{currie2020}
C.~S. Currie, J.~W. Fowler, K.~Kotiadis, T.~Monks, B.~S. Onggo, D.~A.
  Robertson, and A.~A. Tako.
\newblock How simulation modelling can help reduce the impact of covid-19.
\newblock \emph{Journal of Simulation}, 14\penalty0 (2):\penalty0 83--97, 2020.

\bibitem[Foster et~al.(2019)Foster, Jankowiak, Bingham, Horsfall, Teh,
  Rainforth, and Goodman]{foster_variational_2019}
A.~Foster, M.~Jankowiak, E.~Bingham, P.~Horsfall, Y.~W. Teh, T.~Rainforth, and
  N.~Goodman.
\newblock Variational {Bayesian} {Optimal} {Experimental} {Design}.
\newblock In \emph{Advances in {Neural} {Information} {Processing} {Systems}},
  2019.

\bibitem[Jumper et~al.(2021)Jumper, Evans, Pritzel, Green, Figurnov,
  Ronneberger, Tunyasuvunakool, Bates, {\v{Z}}{\'\i}dek, Potapenko,
  et~al.]{jumper2021highly}
J.~Jumper, R.~Evans, A.~Pritzel, T.~Green, M.~Figurnov, O.~Ronneberger,
  K.~Tunyasuvunakool, R.~Bates, A.~{\v{Z}}{\'\i}dek, A.~Potapenko, et~al.
\newblock Highly accurate protein structure prediction with alphafold.
\newblock \emph{Nature}, pages 1--11, 2021.

\bibitem[Kleinegesse and Gutmann(2020)]{kleinegesse_bayesian_2020}
S.~Kleinegesse and M.~U. Gutmann.
\newblock {B}ayesian {Experimental} {Design} for {Implicit} {Models} by
  {Mutual} {Information} {Neural} {Estimation}.
\newblock In \emph{Proceedings of the 37th {International} {Conference} on
  {Machine} {Learning}}, Proceedings of {Machine} {Learning} {Research}. PMLR,
  2020.

\bibitem[Kleinegesse et~al.(2020)Kleinegesse, Drovandi, and
  Gutmann]{kleinegesse2020sequential}
S.~Kleinegesse, C.~Drovandi, and M.~U. Gutmann.
\newblock Sequential {B}ayesian experimental design for implicit models via
  mutual information.
\newblock \emph{{B}ayesian Analysis}, 2020.

\bibitem[Lee et~al.(2011)Lee, Zhang, Munro, and
  Steyvers]{lee_psychological_2011}
M.~D. Lee, S.~Zhang, M.~Munro, and M.~Steyvers.
\newblock Psychological models of human and optimal performance in bandit
  problems.
\newblock \emph{Cog. Sys. Research}, 12\penalty0 (2), 2011.

\bibitem[Myung and Pitt(2009)]{myung_optimal_2009}
J.~I. Myung and M.~A. Pitt.
\newblock Optimal experimental design for model discrimination.
\newblock \emph{Psych. Review}, 116\penalty0 (3), 2009.

\bibitem[Ouyang et~al.(2018)Ouyang, Tessler, Ly, and
  Goodman]{ouyang_webppl-oed_2018}
L.~Ouyang, M.~H. Tessler, D.~Ly, and N.~D. Goodman.
\newblock webppl-oed: {A} practical optimal experiment design system.
\newblock In \emph{Proceedings of the annual meeting of the cognitive science
  society}, 2018.

\bibitem[Overstall and McGree(2020)]{overstall2020}
A.~Overstall and J.~McGree.
\newblock {B}ayesian design of experiments for intractable likelihood models
  using coupled auxiliary models and multivariate emulation.
\newblock \emph{Bayesian Anal.}, 15\penalty0 (1), 2020.

\bibitem[Poole et~al.(2019)Poole, Ozair, Van Den~Oord, Alemi, and
  Tucker]{poole2019}
B.~Poole, S.~Ozair, A.~Van Den~Oord, A.~Alemi, and G.~Tucker.
\newblock On variational bounds of mutual information.
\newblock In \emph{Proceedings of the 36th International Conference on Machine
  Learning}, volume~97 of \emph{Proceedings of Machine Learning Research},
  pages 5171--5180. PMLR, 09--15 Jun 2019.

\bibitem[Robbins(1952)]{robbins_aspects_1952}
H.~Robbins.
\newblock Some aspects of the sequential design of experiments.
\newblock \emph{Bull. Amer. Math. Soc.}, 58\penalty0 (5), 09 1952.

\bibitem[Runge et~al.(2019)Runge, Bathiany, Bollt, Camps-Valls, Coumou, Deyle,
  Glymour, Kretschmer, Mahecha, Mu{\~n}oz-Mar{\'\i},
  et~al.]{runge2019inferring}
J.~Runge, S.~Bathiany, E.~Bollt, G.~Camps-Valls, D.~Coumou, E.~Deyle,
  C.~Glymour, M.~Kretschmer, M.~D. Mahecha, J.~Mu{\~n}oz-Mar{\'\i}, et~al.
\newblock Inferring causation from time series in earth system sciences.
\newblock \emph{Nature communications}, 10\penalty0 (1):\penalty0 1--13, 2019.

\bibitem[Russo et~al.(2017)Russo, Van~Roy, Kazerouni, Osband, and
  Wen]{russo2017tutorial}
D.~Russo, B.~Van~Roy, A.~Kazerouni, I.~Osband, and Z.~Wen.
\newblock A tutorial on thompson sampling.
\newblock \emph{arXiv preprint arXiv:1707.02038}, 2017.

\bibitem[Schulz et~al.(2020)Schulz, Franklin, and Gershman]{schulz2020finding}
E.~Schulz, N.~T. Franklin, and S.~J. Gershman.
\newblock Finding structure in multi-armed bandits.
\newblock \emph{Cognitive psychology}, 119:\penalty0 101261, 2020.

\bibitem[Shahriari et~al.(2015)Shahriari, Swersky, Wang, Adams, and
  de~Freitas]{shahriari_taking_2015}
B.~Shahriari, K.~Swersky, Z.~Wang, R.~P. Adams, and N.~de~Freitas.
\newblock Taking the {Human} {Out} of the {Loop}: {A} {Review} of {Bayesian}
  {Optimization}.
\newblock In \emph{Proceedings of the {IEEE}}, 2015.

\bibitem[Steyvers et~al.(2009)Steyvers, Lee, and
  Wagenmakers]{steyvers_bayesian_2009}
M.~Steyvers, M.~D. Lee, and E.-J. Wagenmakers.
\newblock A {Bayesian} analysis of human decision-making on bandit problems.
\newblock \emph{Journal of Mathematical Psychology}, 53\penalty0 (3), 2009.

\bibitem[Sutton and Barto(2018)]{sutton_reinforcement_2018}
R.~S. Sutton and A.~G. Barto.
\newblock \emph{Reinforcement learning: an introduction}.
\newblock Adaptive computation and machine learning series. The MIT Press, 2nd
  edition, 2018.

\bibitem[Thompson(1933)]{thompson1933likelihood}
W.~R. Thompson.
\newblock On the likelihood that one unknown probability exceeds another in
  view of the evidence of two samples.
\newblock \emph{Biometrika}, 25\penalty0 (3/4):\penalty0 285--294, 1933.

\bibitem[Zhang and Lee(2010)]{zhang_optimal_2010}
S.~Zhang and M.~D. Lee.
\newblock Optimal experimental design for a class of bandit problems.
\newblock \emph{Journal of Mathematical Psychology}, 54\penalty0 (6), 2010.

\end{thebibliography}
}
\clearpage
\section{Appendix}
\subsection{Computational models}
\subsubsection{Win-Stay Lose-Thompson-Sample (WSLTS)}
WSLTS performs Thompson Sampling using the reshaped posterior, excluding the previously selected arm.
That is, WSLTS follows standard Thompson Sampling for Bernoulli bandits, as described in~\citep{russo2017tutorial}, with two important differences:
First, as opposed to sampling from the posterior over all arms, the sampled reward probability of the previously selected arm is set to zero. 
This is to ensure that WSLTS follows the core semantics of WSLS as a strict generalization that allows for more sophisticated exploration/exploitation mechanisms. 
Second, we include a temperature parameter $\lambda$ that is used to reshape the posterior over reward probabilities, to widen or sharpen the posterior, conceptually following~\citet{chapelle2011empirical}. 
That is, we draw rewards for the $k$-th arm from $\textrm{Beta}(\alpha_k^{\lambda},\beta_k^{\lambda})$. 
We treat the model parameter $\lambda$ as a random variable with a $\text{log-normal}(0,1)$ prior distribution. 
The WSLTS model has three parameters: $\gamma_w$, which controls the probability of staying after winning, $\gamma_l$, which controls the probability of shifting after losing, and $\lambda$, for posterior reshaping.

\subsubsection{Autoregressive $\varepsilon$-Greedy (AEG)}
The chance of selecting the previously selected arm is controlled by the $\varphi$ parameter. 
That is, the probability of selecting the previous arm is given by $\varphi + \frac{1-\varphi}{m}$, where on exploration choices, $m$ is the number of bandit arms, and on greedy choices, $m$ is the number of arms that have the maximal expected reward probability, thereby breaking possible ties.
$\varphi$ can thus be thought of as a ``stickiness'' parameter~\citep{schulz2020finding} that favors the previous arm on both greedy as well as exploration choices.
The AEG model thus has two parameters, $\varepsilon$ and $\varphi$.

\subsubsection{Generalized Latent State (GLS)}
\citet{lee_psychological_2011} proposed a latent state model for bandit tasks, which includes a latent \emph{explore}/\emph{exploit} state $z_i$ for each trial $i$. 
This latent state determines which arm is chosen to resolve the explore-exploit dilemma whenever one arm has more observed wins but also more losses than the other arm(s). 
Their model proposes that the latent state changes stochastically on a trial-by-trial basis, where transitions between latent states are modeled using a $\textrm{Bernoulli}(0.5)$ distribution.

The authors demonstrate that this model captures people's behavior in an empirical evaluation, but they also show that most participants could be modeled by a simplified account, in which there is only one switch point from exploration to exploitation, without the possibility of transitioning back to exploration~\citep[for details, see][]{lee_psychological_2011}.

The originally proposed latent state model by~\citep{lee_psychological_2011} only dealt with 2-armed bandits.
Here we describe the generalization used for k-armed bandits.~\footnote{We thank Patrick Laverty for contributing towards this.}
We distinguish between the following situations the agent can be in~\citep{lee_psychological_2011}.
\begin{description}
\item[Same] If two or more arms have the maximum number of wins and the minimum number of losses, choose one of these at random. 

\item[Better-worse] If only one arm has the maximum number of wins and the minimum number of losses, choose this arm. 

\item[Explore-exploit] If neither \textit{same} nor \textit{better-worse} applies, the agent faces an exploration-exploitation dilemma, which is resolved based on the latent state of the agent as follows:
\begin{description}
\item[Exploit state] If the latent state is \textit{exploit}, then, if there is at least one arm with more wins than all other arms, choose the arm with the maximum number of wins that has the minimum number of losses out the set of arms that have the maximum number of wins (or choose uniformly at random from the set of equivalent options). 
\item[Explore state] If the latent state is \textit{explore}, then, if there is at least one arm with fewer losses than all other arms, choose the arm with the maximum number of wins out of the set of arms with the minimum number of failures (or choose uniformly at random from the set of equivalent options). 
\end{description}
\end{description}

We propose a simple extension of this model, which we call the generalized latent state (GLS) model and that includes the original latent state model and the kind of switch-point model as special cases, but also accommodates more flexible latent state transitions. 
The GLS models the probability of being in a latent exploit (as opposed to explore) state as being dependent on whether the last latent state was explore or exploit, and on whether a reward was observed in the previous trial or not.

The generalization to dependencies in the latent transitions allows the model to account for several further psychologically plausible mechanisms that may be at play.
First, as suggested by the switch-point model, people may have a ``stickiness'' (or ``anti-stickiness'') in their latent state, such that the probability of being in an exploit-state at trial $t$ depends on the previous latent state.
Note that the stickiness in the latent state may be different for explore as opposed to exploit states, and we therefore include these two transition probabilities as two separate model parameters.
Second, switches between the latent explore/exploit state may not be symmetric in regard to the previously observed reward.
That is, people may, e.g., be more willing to switch from exploration to exploitation when observing a win than when observing a loss, as would be suggested by the WSLS heuristic. The transition probabilities when observing a win and when observing a loss are thus also treated as two separate model parameters.

In total, the GLS has an accuracy of execution parameter $\gamma_{\textrm{GLS}}$ (conceptually identical to the one in the latent state model of~\citet{lee_psychological_2011}) which accounts for random behavior, as well as four parameters controlling the latent state transitions described above.
The initial latent state is sampled from a $\text{Bernoulli}(0.5)$ distribution for simplicity. 
Note that our proposed GLS model can recover the latent state model by \citep{lee_psychological_2011} by setting all latent transition probabilities to $0.5$, rendering the latent state independent of the preceding state and reward. 
Similarly, we could treat the latent exploitation state as absorbing, such that once a transition to the exploit state has happened, the agent never goes back to exploring. 

\subsection{MINEBED method}
In their MINEBED method,~\citep{kleinegesse_bayesian_2020} proposed to maximize a cheaper, tractable lower bound on the MI, as opposed to spending resources on estimating the MI to a high accuracy. This lower bound is parametrized by a neural network $T_{\psib}(\bm{v}, \ybf)$, where $\psib$ are the neural network parameters.\footnote{This neural network takes as input $\bm{v}$ and $\ybf$, which are concatenated, and returns a scalar value.} The MINEBED method uses the NWJ lower bound~\citep[see e.g.][]{poole2019}, which is now a function of $\dbf$ and $\psib$, given by
\begin{equation} \label{eq:lb}
    U(\dbf; \psib) = \mathbb{E}_{p(\ybf|\bm{v},\dbf)p(\bm{v})}\left[T_{\psib}(\bm{v}, \ybf)\right] - e^{-1} \mathbb{E}_{p(\ybf|\dbf)p(\bm{v})}\left[e^{T_{\psib}(\bm{v}, \ybf)}\right] \\
\end{equation}
The expectations in Equation~\ref{eq:lb} are approximated via sample averages.

The lower bound shown in Equation~\ref{eq:lb} is a function of the experimental designs $\dbf$ and the neural network parameters $\psib$. For a fixed design, we can tighten the lower bound, i.e.~let it approach the true MI value, by optimizing the neural network with respect to $\psib$ by means of stochastic gradient-ascent, using Equation~\ref{eq:lb} as an objective function. The gradients of $U(\dbf; \psib)$ with respect to $\psib$ can be easily obtained via automatic differentiation available in common machine-learning libraries. This allows us to obtain an estimate of the mutual information at a fixed design, i.e.
\begin{equation}
    U(\dbf) = \max_{\psib} U(\dbf; \psib)
\end{equation}

In order to then optimize the MI with respect to the designs $\dbf$ we can use any gradient-free optimization technique. In our case, we use Bayesian optimization (BO)~\citep{shahriari_taking_2015}. As a probabilistic surrogate model we use Gaussian Processes (GPs) and as an acquisition function we use Expected Improvement (EI). We present a short summary of the gradient-free MINEBED method in Algorithm~\ref{algo:minebed}. Note that a variant of the MINEBED method deals with gradient-based design optimization~\citep[see][]{kleinegesse_bayesian_2020}. However, this is not applicable in our experiments with bandit tasks, where gradients with respect to the discrete choices and rewards $\ybf$ are undefined.
\begin{algorithm}[!t]
\caption{Gradient-Free MINEBED}\label{algo:minebed}
\begin{algorithmic}[1]
\Input Implicit simulator model $\ybf \sim p(\ybf|\bm{v}, \dbf)$, prior distribution $p(\bm{v})$, neural network architecture for $T_{\psib}(\bm{v}, \ybf)$
\Output Optimal design $\dbf^\ast$, trained neural network $T_{\psib^{\ast}}(\bm{v}, \ybf)$ at $\dbf^\ast$
\item[]
\State Randomly initialize the experimental designs $\dbf_n \leftarrow \dbf_0$
\State Initialize the Gaussian Process for BO
\While {$U(\dbf_n)$ not converged}
    \State {Sample from the prior: $\bm{v}^{(i)} \sim p(\bm{v})$ for $i=1, \dots, N$}
    \State {Sample from the simulator: $\ybf^{(i)} \sim p(\ybf|\bm{v}^{(i)}, \dbf)$ for $i=1, \dots, N$}
    \State {Randomly initialize the neural network parameters $\psib_n \leftarrow \psib_0$}
    \While {$U(\dbf_n; \psib_n)$ with fixed $\dbf_n$ not converged}
        \State {Compute a sample average of the lower bound in Equation~\ref{eq:lb}}
        \State {Estimate gradients of sample average with respect to $\psib_n$}
        \State {Update $\psib_{n}$ using any gradient-based optimizer}
    \EndWhile
    \State {Use $\dbf_n$ and $U(\dbf_n)$ to update the Gaussian Process}
    \State {Use BO to find out at which $\dbf_{n+1}$ to evaluate next}
\EndWhile
\end{algorithmic}
\end{algorithm}
\paragraph{Posterior Estimation} By looking at Equation~\ref{eq:lb}, we can see that the NWJ lower bound is tight when $T_{\psib^{\ast}}(\bm{v}, \ybf) = 1 + \log p(\bm{v}|\ybf, \dbf^\ast) / p(\bm{v})$. By rearranging this, we can thus use our trained neural network $T_{\psib^{\ast}}(\bm{v}, \ybf)$ to compute a (normalized) estimate of the posterior distribution,
\begin{equation}
    p(\bm{v}|\ybf, \dbf^\ast) = p(\bm{v}) e^{T_{\psib^{\ast}}(\bm{v}, \ybf) - 1}.
\end{equation}

\subsection{Experimental Details}

We here provide details about our experiments.

\paragraph{Priors} We use a uniform categorical prior over the model indicator $m$, i.e.~$p(m) = \mathcal{U}(\{1, 2, 3\})$. We generally use uninformative priors $\mathcal{U}(0, 1)$ for all model parameters, except for the temperature parameter of the WSLTS model that has a $\text{LogNorm}(0, 1)$ prior, as it acts as an exponent in reshaping the posterior. We generate $50{,}000$ samples from the prior and then simulate corresponding synthetic data $\ybf | \thetab, \dbf$ at every design $\dbf$.

\paragraph{Sub-networks} For all of our experiments, we use sub-networks $\mathbf{S}_{\psib}(\bm{v}, \ybf)$ that consist of two hidden layers with 64 and 32 hidden units, respectively, and ReLU activation functions. The number of sufficient statistics we wish to learn for each block of behavioral data is given by the number of dimensions in the output layer of the sub-networks. These are $6$, $8$, $6$ and $8$ units for the MD, PE WSLTS, PE AEG and PE GLS experiments, respectively. The flexibility of a sub-network is naturally increased when increasing the number of desired summary statistics, but the cost increases accordingly. When the number of summary statistics is too low, the summary statistics we learn may not be sufficient. We have found the above number of summary statistics to be effective middle-grounds.

\paragraph{Main Network} The main network $T_{\psib}(\bm{v}, \ybf)$ consists of the concatenated outputs of the sub-networks for each block of behavioral data and the variable of interest. This is then followed by two fully-connected layers with ReLU activation functions. For the MD experiment we use $32$ hidden units for the two hidden layers, while we use $64$ and $32$ hidden units for the PE experiments. See Figure~\ref{fig:my_label} for a visualization of this bespoke neural network architecture.

\paragraph{Training} We use the Adam optimizer to maximize the lower bound shown in Equation~\ref{eq:lb}, with a learning rate of $10^{-3}$ and a weight decay of $10^{-3}$ (except for the PE WSLTS experiments where we use a weight decay of $10^{-4}$). We additionally use a plateau learning rate scheduler with a decay factor of $0.5$ and a patience of $25$ epochs. We train the neural network for $200$, $400$, $300$ and $300$ epochs for the MD, PE WSLTS, PE AEG and PE GLS experiments, respectively. At every design we simulate $50{,}000$ samples from the data-generating distribution (one for every prior sample) and randomly hold out $10{,}000$ of those as a validation set, which are then used to compute an estimate of the mutual information via Equation~\ref{eq:lb}. During the BO procedure we select an experimental budget of $400$ $U(\dbf)$ evaluations ($80$ of which were initial evaluations), which is more than double needed to converge.


\subsection{Additional Results}

\begin{figure}[!t]
    \centering
    \includegraphics[width=0.7\linewidth]{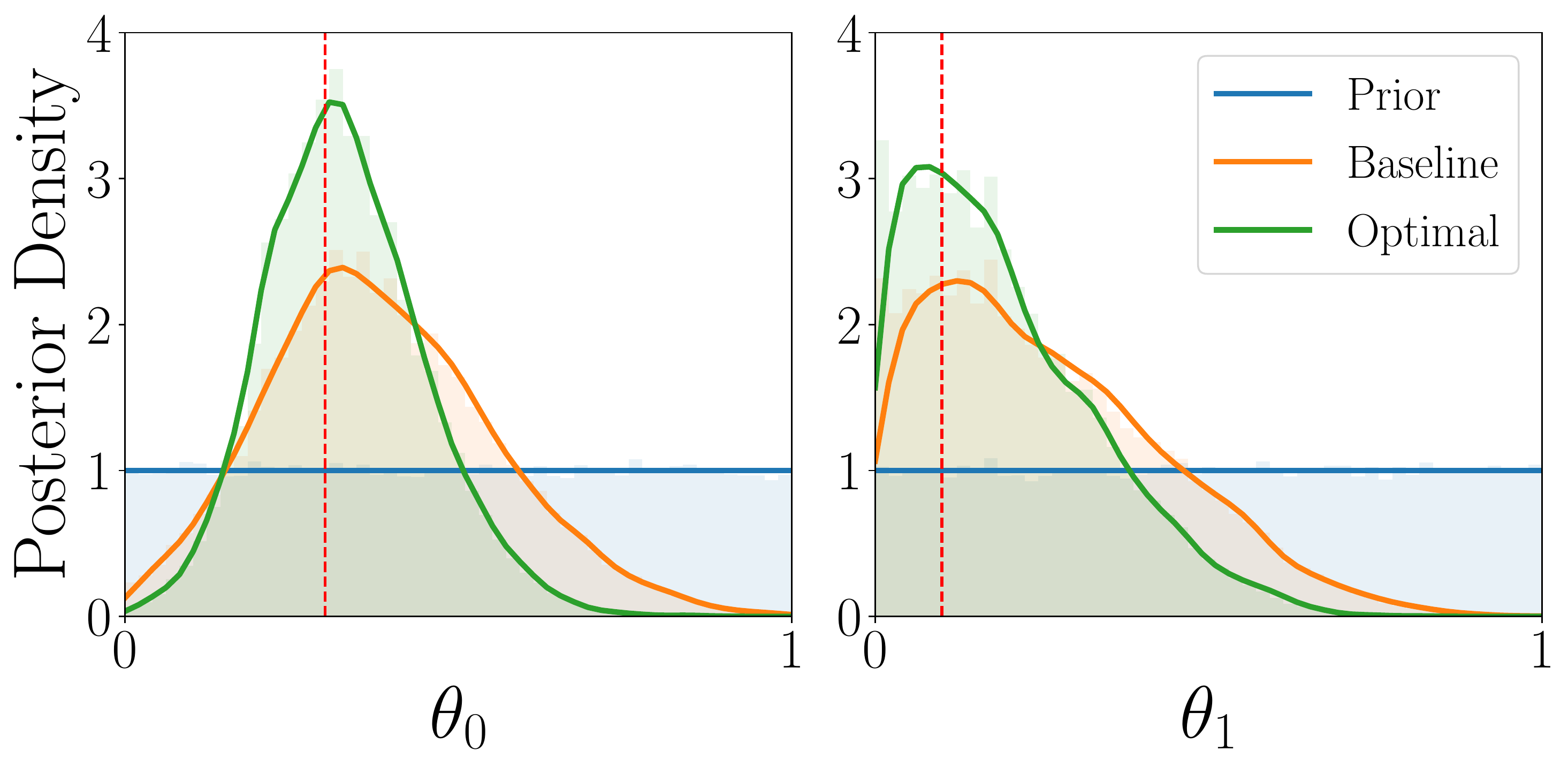}
    \vspace{0.1cm}
        \caption{Marginal posterior distributions of the two AEG model parameters for optimal (green) and baseline (orange) designs, averaged over $1{,}000$ observations.}
        \label{fig:pe_post_aeg}
\end{figure}

\begin{figure}[!t]
    \centering
    \includegraphics[width=1\linewidth]{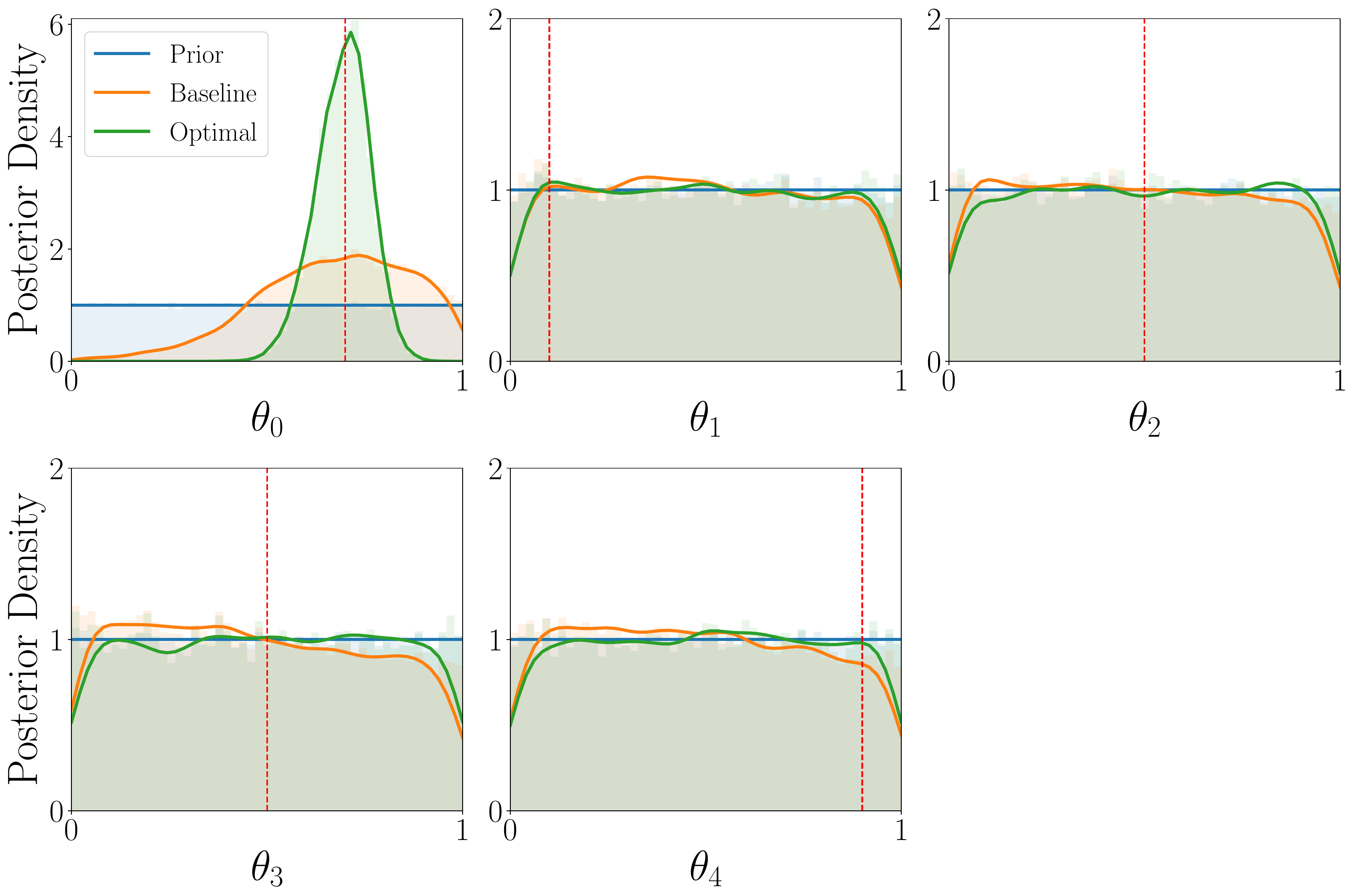}
    \vspace{0.1cm}
        \caption{Marginal posterior distributions of the five GLS model parameters for optimal (green) and baseline (orange) designs, averaged over $1{,}000$ observations. Note that the downwards behavior at the boundaries is due to taking a Gaussian kernel density estimation of a nearly-uniform distribution.}
        \label{fig:pe_post_gls}
\end{figure}

We here present additional parameter estimation results for the AEG in Figure~\ref{fig:pe_post_aeg} and for the GLS model in Figure~\ref{fig:pe_post_gls}.

Similarly to the parameter estimation task of the WSLTS model (see Figure~\ref{fig:pe_post}), our optimal designs yield data that are more informative than baseline designs from literature, which can be seen from sharper posterior distributions that are centered around the ground-truths (shown in red-dotted lines). We note that the transition parameters of the GLS model are not well-estimated for any of the designs due to the uninformative priors and small number of trials in the behavioral blocks.

\end{document}